\documentclass[journal]{IEEEtran}

\ifCLASSINFOpdf
  \usepackage[pdftex]{graphicx}
  \graphicspath{{../pdf/}{../jpeg/}}
  \DeclareGraphicsExtensions{.pdf,.jpeg,.png}
\else
  \usepackage[dvips]{graphicx}
  \graphicspath{{../eps/}}
  \DeclareGraphicsExtensions{.eps}
\fi
\usepackage{diagbox}             
\usepackage{multirow}            
\usepackage{graphicx}            
\usepackage[subfigure]{graphfig} 
\usepackage{bm}
\usepackage{stfloats}
\usepackage{lineno,hyperref}
\modulolinenumbers[5]
\usepackage{booktabs}
\usepackage[ruled]{algorithm2e}
\usepackage{bm}
\usepackage{algorithmic} 
\usepackage{xcolor}

\usepackage{amsmath}

\usepackage{subfigure}

\begin{document}

%
\title{Embedded Self-Distillation in Compact Multi-Branch Ensemble Network for Remote Sensing Scene Classification}

\author{Qi Zhao,~\IEEEmembership{Member,~IEEE,}
        Yujing Ma, Shuchang Lyu, Lijiang Chen
\thanks{Qi Zhao is with the Department of Electronics and Information Engineering, Beihang University, Beijing, 100191, China, e-mail: zhaoqi@buaa.edu.cn.}
\thanks{Yujing Ma is with the Department of Electronics and Information Engineering, Beihang University, Beijing, 100191, China, e-mail: zy1902407@buaa.edu.cn.}
\thanks{Shuchang Lyu is with the Department of Electronics and Information Engineering, Beihang University, Beijing, 100191, China, e-mail: lyushuchang@buaa.edu.cn.}
\thanks{Lijiang Chen is with the Department of Electronics and Information Engineering, Beihang University, Beijing, 100191, China, e-mail: chenlijiang@buaa.edu.cn.}
}

\markboth{Journal of \LaTeX\ Class Files,~Vol.~14, No.~8, August~2015}%
{Shell \MakeLowercase{\textit{et al.}}: Bare Demo of IEEEtran.cls for IEEE Journals}

\maketitle

\begin{abstract}
Remote sensing (RS) image scene classification task faces many challenges due to the interference from different characteristics of different geographical elements. To solve this problem, we propose a multi-branch ensemble network to enhance the feature representation ability by fusing features in final output logits and intermediate feature maps. However, simply adding branches will increase the complexity of models and decline the inference efficiency. On this issue, we embed self-distillation (SD) method to transfer knowledge from ensemble network to main-branch in it. Through optimizing with SD, main-branch will have close performance as ensemble network. During inference, we can cut other branches to simplify the whole model. In this paper, we first design compact multi-branch ensemble network, which can be trained in an end-to-end manner. Then, we insert SD method on output logits and feature maps. Compared to previous methods, our proposed architecture (ESD-MBENet) performs strongly on classification accuracy with compact design.  Extensive experiments are applied on three benchmark RS datasets AID, NWPU-RESISC45 and UC-Merced with three classic baseline models, VGG16, ResNet50 and DenseNet121. Results prove that our proposed ESD-MBENet can achieve better accuracy than previous state-of-the-art (SOTA) complex models. Moreover, abundant visualization analysis make our method more convincing and interpretable.
\end{abstract}


%
\IEEEpeerreviewmaketitle
\section{Introduction}
\IEEEPARstart{R}{emote} sensing scene classification is a recent popular task in practical application. It reveals the geographical characteristic, such as land utilization, vegetation coverage~\cite{Land_utilize}. With the progress of RS scene classification, researches on local land planning, tree planting and afforestation can be realized more intelligent. In recent years, with the rapid development of deep learning technology~\cite{deep_learning_technology, deep_learning_technology1}, methods for improving the RS scene classification accuracy have been continuously proposed. 

\begin{figure}[ht]
\centering
\includegraphics[width=1.0\linewidth]{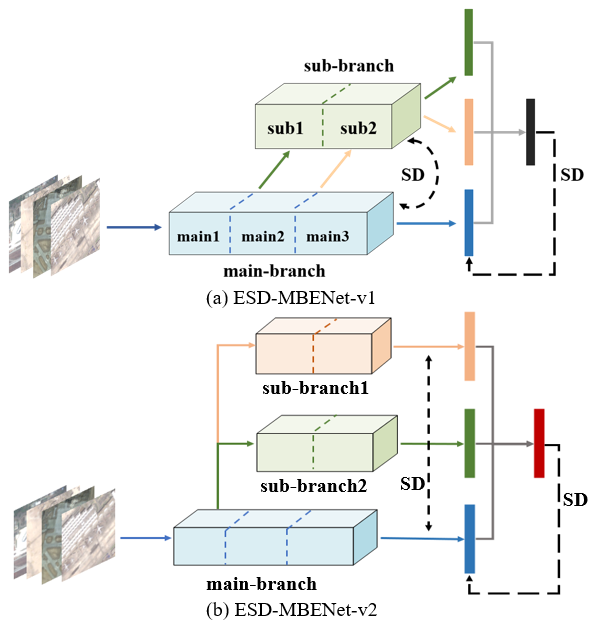}
\caption{The overview of ESD-MBENet structure. ``SD'' denotes self-distillation. (a) ESD-MBENet-v1, (b) ESD-MBENet-v2.}
\label{Fig1}
\end{figure}

\par In RS scene classification task, RS images always have large resolution and large geographic coverage area. Therefore, the major problem is the interference from different characteristics of different geographical elements. To solve this problem, previous works focus on fusing multi-level features to enhance the model's ability of representing complex structural information. In this paper, we integrate ensemble learning method into CNN modules to construct a multi-branch ensemble network. Towards different geographical elements in RS images, we use more branches to provide sufficient representation. Through fusing the final logits of different branches, we combine all perspectives together and obtain a more convincing prediction. As shown in Fig.~\ref{Fig1}, we use main-branch and sub-branch to construct our ensemble network.
\par Although ensemble network is quite effective on dealing with the above-mentioned problem, the large memory and computation cost of multi-branch structure can not be ignored. Especially on embedded device like UAV, multi-branch models are cumbersome and hard to deploy. To construct a lighter yet high-efficient multi-branch network, we embed self-distillation (SD) method in multi-branch ensemble network (ESD-MBENet). The overview of ESD-MBENet is shown in Fig.~\ref{Fig1}. 
\par To intuitively lighten the multi-branch structure, we design one sub-branch in ESD-MBENet-v1. Then we split the two branches and connect blocks in a zigzag manner. Designing like this, we can generate a multi-logits network with only two branches. As shown in Fig~\ref{Fig1}(a), we respectively split the main-branch into three blocks and sub-branch into two blocks. The images are first fed into the first block of main-branch (``main1''). The output feature maps then respectively pass through ``sub1-sub2'' and ``main2''. The output feature maps of ``main2'' will then pass through ``sub2'' and ``main3'' respectively. Finally, we obtain three output logits from three paths (``main1-main2-main3'', ``main1-sub1-sub2'', ``main1-main2-sub2''). If we set more split points, we can get more diverse output logits. Besides ESD-MBENet-v1, we also design ESD-MBENet-v2 as Fig.\ref{Fig1}(b) to form the multi-branch ensemble network because if we set split points in more deeper layer, the number of branches of ESD-MBENet-v1 will be reduced. ESD-MBENet-v2 can flexibly construct multiple branches without being affected by the backward movement of the split points. Essentially, we construct weight-sharing blocks in ESD-MBENet and maximally explore the representation ability of them.
\par Even though we use weight-sharing blocks to simplify multi-branch network, sub-branch is also cumbersome during inference. To cast off sub-branch during inference, we introduce self-distillation method to transfer knowledge from multi-branch ensemble network to the main-branch. When main-branch can show comparable performance as the ensemble network, we can prune the sub-branch and only adopt main-branch during inference. In ESD-MBENet, we embed SD method into the ensemble network. Specifically, the final ensemble logits which are fused together by logits of every branch will be served as soft-label to distill knowledge to main-branch. In intermediate feature maps, the ensemble feature map is used to guide feature map of main branch. After optimizing with SD method, main-branch has close performance as ensemble network. Therefore, we can prune sub-branch and only use main-branch as inference model. 

\par Compared to previous single-branch networks~\cite{single-branch1,single-branch2,single-branch3}, the inference speed of our proposed model does not become slower. Compared to previous multi-branch networks~\cite{Multi-branch1,multi-branch2,multi-branch3,Multi-Granularity}, our proposed ESD-MBENet achieves better performance with more compact structure by enhancing the capability of main-branch to learn more and better knowledge. Extensive experiments on using VGG16~\cite{VGG}, ResNet50~\cite{ResNet} and DenseNet121~\cite{densenet} as baseline models on RS benchmark datasets (AID~\cite{AID}, NWPU-RESISC45~\cite{NWPU} and UC-Merced~\cite{UC-Merced}) prove the effectiveness of our proposed ESD-MBENet. Classification results on ESD-MBENet surpass previous method and reach SOTA level. To show generalization of our model, we also conduct experiments on natural scene classification datasets (CIFAR and tiny-ImageNet). The results are also encouraging. Our main contributions can be summarized as follows:
\begin{itemize}
\item We propose an compact yet efficient multi-branch ensemble network embedded with self-distillation method in RS scene classification to overcome the interference of different geographical elements in RS images.
\item We insert self-distillation method in ensemble network to distill knowledge to main-branch, which can further simplify the whole network.
\item Our proposed ensemble networks and main-branch network all achieve better classification results than previous SOTA networks on RS image datasets and natural scene datasets.
\end{itemize}
 

\section{RELATED WORKS}
\subsection{RS Image Classification}
With the development of artificial intelligence, RS scene classification methods have transitioned from handcrafted feature extraction to deep learning feature extraction.
\par Handcrafted feature extraction is first applied in RS image classification tasks. Typical handcrafted feature extraction methods are SIFT~\cite{SIFT,SIFT1,SIFT_2}, HOG~\cite{HoG_RS,HOG1,HOG2,HOG3} and etc. Low-level information of RS images can be extracted by these methods. \cite{BIC1,BIC2,BIC3} propose the border-interior pixel classification (BIC), which can calculate the border and interior pixels color histograms. Later, principal component analysis (PCA),  K-means clustering, bag-of-visual-words (BoVW)~\cite{UC-Merced, 2-D_BOVW, BoVW_RS, BoVW_study} and sparsely encoding~\cite{sparse-encoding} 
are proposed to extract mid-level information of the images for RS scene classification. 
\par Due to the appearance of back-propagation neural network, deep learning has developed quickly. \cite{AlexNet,VGG,ResNet,densenet} achieve amazing improvement in image classification task. Based on these baseline models, RS image classification technologies have improved rapidly~\cite{RSBD,RSBD1,RSBD2}. \cite{DLBFS} proposes a deep-learning-based feature-selection method to achieve feature abstraction of the RS images.~\cite{DLBCRSI} classifies unlabeled RS images, which improves the speed and accuracy of classification compared to traditional machine learning algorithms. Our proposed ESD-MBENet also uses deep learning method for RS scene classification.
\subsection{Multi-Branch Network}
A multi-branch network can obtain abundant information from multiple perspectives of the input images, which helps the network to have a more comprehensive representation of the images and improves the generalization of the classifier. In RS scene classification task, many researchers explore the potential of multi-branch networks by fusing features of different branches~\cite{multi-branch6,multi-branch7,multi-branch8}. ~\cite{RS-MSSF,multi-branch4,multi-branch5} achieve feature fusion method by extracting multiple spectral and spatial features and concatenate them, which improves the accuracy of RS image classification. ~\cite{Multi-branch1} uses a multi-branch lightweight network to extract image features, and builds a graph model based on the learned features. ~\cite{multi-branch2} adopts fine-grained and coarse branch to obtain the features in images. 
\par Our proposed ESD-MBENet uses fewer modules and more weight-sharing blocks to build a multi-branch network, which can obtain multi-view information from multi-branch, so that the network can have more references when making final decisions.
\begin{figure*}
  \centering
  \includegraphics[width=1.0\linewidth]{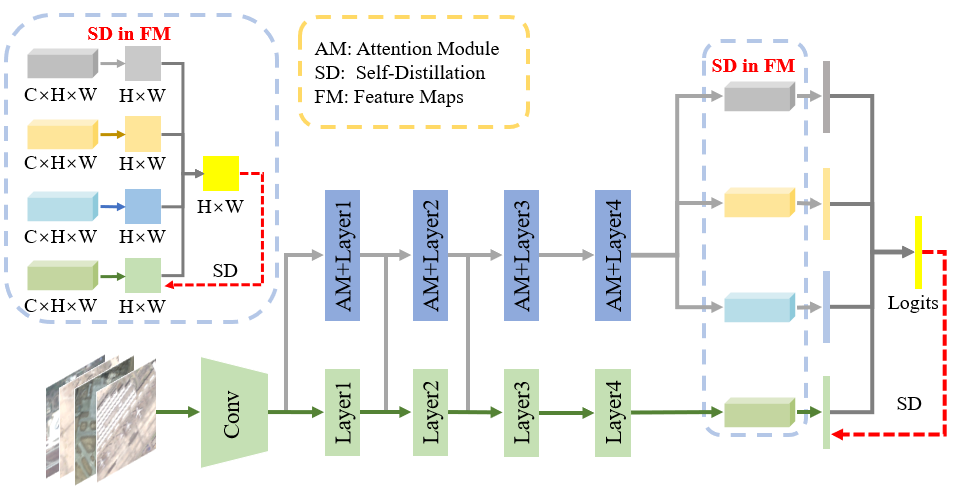}
  \caption{Overall framework of the ESD-MBENet-v1. The bottom branch is the main-branch. The sub-branch and the main-branch use the same backbone, namely VGG16 or ResNet50 or DenseNet121. Compared to the main-branch, the sub-branch adds an attention module, such as SE or CAM or Dropout. It is designed to overcome the interference of different geographic elements in RS images and the complexity of the model. Four branches are used for online training. We self-distill the final output logits and intermediate feature maps. In addition, the hard labels are also used to optimize the network. To reduce the complexity of the model, only the main-branch is used to complete the model inference.}
\label{Fig2}
\end{figure*}
\subsection{Knowledge Distillation and Self-Distillation}
Knowledge distillation is a concept proposed by Hinton~\cite{KD}. The main purpose of knowledge distillation and self-distillation is model compression. Knowledge distillation aims to guide simple and relatively poor student network learning from a complex but superior teacher network~\cite{model-compress1,model-compress2}. The student network learns how the teacher network learns to improve its distinguishing performance for RS image classification. Self-distillation mainly distilled from its own network, without the assistance of external networks or models. The weighted combination of multiple teacher networks is proposed to guide students to learn from it in~\cite{efficient_kd}. A knowledge distillation framework is proposed in~\cite{TSNKD}, which makes the output of the student and teacher models match. Discriminative modality distillation approach is introduced in~\cite{DMD}, the teacher is trained on multimodal data and then the student model learns from the teacher model to improve the performance of the RS image classifiction. To address the problem of network overfitting due to noisy data, a novel noisy label distillation method (NLD) is proposed in~\cite{NLD}. 
\par Regarding the self-distillation method, there is little research in RS image classification task. We propose an end-to-end compact multi-branch ensemble network ESD-MBENet that uses self-distillation to improve the main-branch performance.
\section{PROPOSED NETWORK}
\subsection{Overview of ESD-MBENet}
To overcome the interference of different geographic elements in RS images, we propose two versions of ESD-MBENet. Fig.~\ref{Fig2} shows the structure of ESD-MBENet-v1. For the purpose of using fewer modules and constructing a multi-branch network, we set multiple split points in the network. First, we design a network of two branches, the main-branch and the sub-branch. The main-branch and the sub-branch use the same backbone. To resist the lack of multi-branch diversity caused by partial weight-sharing, attention modules such as SE or CAM or Dropout are added to the sub-branch. If the sub-branch share the first ``conv'' with the main-branch, we can set split points in the main-branch ``layer1'' and ``layer2'', and then send the main-branch feature maps to the sub-branch network in the position of the split points. Of course, the main-branch and the sub-branch can also share the ``layer1'', and the corresponding split points will move backwards. In the training phase, multi-branch output results can be obtained after fusing the output logits. To speed up the inference speed and simplify the model complexity, we only use the main-branch for inference.
\par
\begin{figure*}
  \centering
  \includegraphics[width=1.0\linewidth]{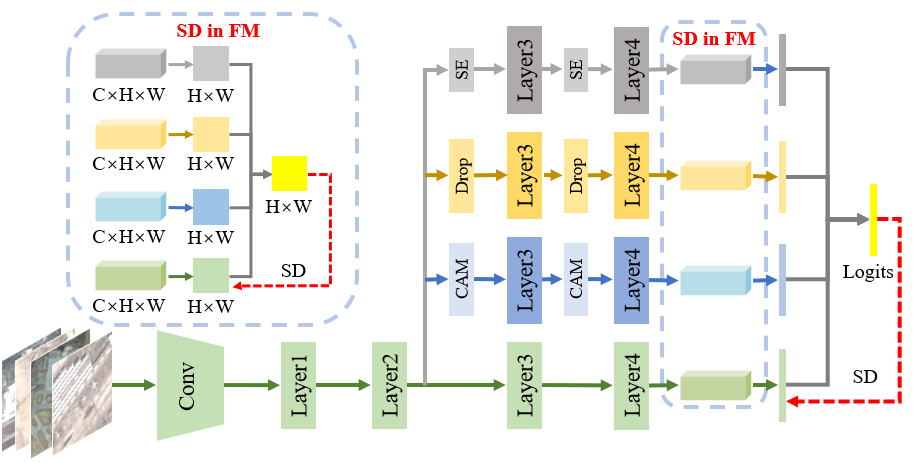}
  \caption{The structure of the ESD-MBENet-v2. When the split points keep moving backwards in ESD-MBENet-v1, the number of branches decreases. In order to maintain the diversity of the network, new branches are added as compensation. To ensure the diversity of feature extraction under the premise of more weight-sharing, different attention modules are added to each branch, namely SE, CAM and Dropout. The process of training optimization and inference are the same as ESD-MBENet-v1.}
\label{Fig3}
\end{figure*}

 As the split points move backwards, more and more weights are shared by the network, and the number of multiple branches that can be constructed is reduced. It will bring about the lack of network diversity and the deterioration of network performance. To solve the problem, while sharing as many weights as possible to build multi-branch network, we propose ESD-MBENet-v2. The network structure diagram is shown in Fig.~\ref{Fig3}. ESD-MBENet-v2 has four branches. The same backbone is used in four branches, and different attention modules are added respectively to each sub-branch, namely SE, CAM, and Dropout. There is no weight-sharing in each branch. ESD-MBENet-v2 is not limited by the number of branches. At the corresponding split point, we can add multiple branches at will. Taking into account the amount of parameters, model complexity and performance improvement, we choose four branches when exprimenting. Similar to ESD-MBENet-v1, multiple branches are used for training, and the main-branch is used as inference prediction.
\par In short, ESD-MBENet-v1 uses fewer modules to construct multi-branch structures, and ESD-MBENet-v2 can construct multi-branch networks more flexibly to realize multi-branch networks. Both ESD-MBENet-v1 and ESD-MBENet-v2 consider using as few modules as possible to build multiple branches, that is, sharing as many weights as possible. Through experimental verification, the two versions of ESD-MBENet we proposed both have better RS image classification performance than previous methods.
\begin{figure}[ht]
\centering
\includegraphics[width=1.0\linewidth]{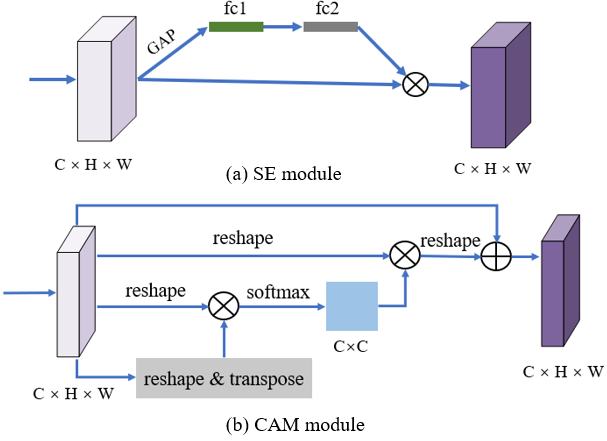}
\caption{Structure diagram of SE and CAM module. (a)SE module: The obtained feature maps (C×H×W) pass through the global average pooling layer and two fully connected layers (C×1×1). Multiply it with the originally obtained feature maps (C×H×W) to obtain new feature maps (C×H×W) that contains richer image information. (b)CAM module: After the obtained feature maps (C×H×W) are reshaped, transposed, multiplied and etc, new feature maps containing the channel attention mechanism is obtained.}
\label{Fig4}
\end{figure}
\subsection{Attention Modules}
The diversity of multi-branch networks is very important for feature fusion. In our proposed ESD-MBENet, if the sub-branch and the main-branch are exactly the same, it may lack diversity for image feature extraction, and it is impossible to describe image features from multiple perspectives. Therefore, we propose the following design, the sub-branch and the main-branch use the same backbone, the main-branch does not add any extras, and the attention module is added to the sub-branch. Compared with tasks such as image segmentation, image classification does not require so much attention between pixels. Therefore, we consider the enhancement of the attention between feature map channels. We use the SE and CAM modules proposed in SENet~\cite{SE} and DANet~\cite{CAM} to add attentions to the feature maps.
\par The structure diagram of SE and CAM modules is shown in Fig.~\ref{Fig4}. The realization idea of SE is to take the obtained feature maps and pass it through the global average pooling layer, two fully connected layers, the sigmoid function as Eq.~\ref{eq1}, and then multiply it with the originally input feature maps. In this way, the global information of the images can be integrated into the feature maps, which improves the sensitivity of the network to the channel, and makes the feature maps contain richer information.
\begin{equation}
    S(x) = \frac{1}{1+e^{-x}}
    \label{eq1}
\end{equation}
\par The CAM module selectively emphasizes the interdependent channel mappings by integrating the relevant features between all channel mappings. After the input feature maps are reshaped, transposed and multiplied, the matrix of C×C is obtained, which is the channel attention map. Then multiply the matrix which should pass through the softmax layer with the input feature maps after reshaping. Finally, reshape the feature maps and add the feature maps with the originally input feature maps. In this way, the attention mechanism is added to the feature maps. The specific process of SE and CAM module is consistent with~\cite{SE} and~\cite{CAM}.
\par In addition, for RS image classification, an image corresponds to a category, but not all pixel values in the image can provide useful information for the classification results, and even some pixels may interfere with the judgment for the image. Therefore, we design to use the Dropout module, and the probability of a random drop is 0.2. This is of great help to the improvement of network generalization performance.
\par These three modules are all independent modules, which can be embedded anywhere in the network without affecting other structure of the network. They have strong flexibility.
\subsection{Multi-Branch Ensemble}
To solve the interference between different geographic elements in RS images, we propose ESD-MBENet. ESD-MBENet-v1 constructs multiple branches from the two branches (main-branch and sub-branch) networks by setting different split points, and then fuses the features in multiple branches. Assume that the main-branch and the sub-branch share the first ``conv'', and we build a total of four branches as Fig.~\ref{Fig2}. The construction of multi-branch is as follows. The first branch is the main-branch, the second branch is the sub-branch. The third branch is the feature maps obtained after ``layer1'' of the main-branch pass through the rest of the sub-branch, and the fourth branch is the feature maps after ``layer2'' of the main-branch pass through the rest of the sub-branch. Suppose the main-branch is divided into ``conv'', ``main-layer1'', ``main-layer2'', ``main-layer3'' and ``main-layer4'', mathematically expressed as ${f_{0}}$,${f_{1}}$,${f_{2}}$,${f_{3}}$,${f_{4}}$, and the sub-branch is divided into ``sub-layer1'', ``sub-layer2'', ``sub-layer3'', ``sub-layer4'', mathematically expressed as ${g_{1}}$,${g_{2}}$,${g_{3}}$,${g_{4}}$. To ensure the data diversity after the multi-branch with weight-sharing, the sub-branch adds the same attention module in front of each layer compared to the main-branch, such as adding SE or CAM or Dropout. The four branches ${b_{1}}$,${b_{2}}$,${b_{3}}$,${b_{4}}$ divided by ESD-MBENet-v1 are  $b_{1} = {f_{4}}{f_{3}}{f_{2}}{f_{1}}{f_{0}}(x), b_{2} = {g_{4}}{g_{3}}{g_{2}}{g_{1}}{f_{0}}(x), b_{3} = {g_{4}}{g_{3}}{g_{2}}{f_{1}}{f_{0}}(x), b_{4} = {g_{4}}{g_{3}}{f_{2}}{f_{1}}{f_{0}}(x)$, which \bm$x$ is a batch of images. The ESD-MBENet-v1 ensemble multi-branch algorithm is shown in Alg.~\ref{alg1}.
\begin{algorithm} 
\caption{ESD-MBENet-v1 multi-branch algorithm} 
\label{alg1} 
\begin{algorithmic}[1] 
\REQUIRE A batch of images \bm{$x$}. Define main-branch blocks function as a list $[f_{0},f_{1},...,f_{m}]$, the sub-branch blocks function as a list $[g_{1},g_{2},...,g_{m}]$, the split points after the corresponding layer as a list $sp=[0,1,2,...,n], n \le m-1$, main-branch fully connected layer function the {$f_{c}$}, {$k^{th}$} sub-branch fully connected layer function {$f_{c_{k}}$}.
\ENSURE the ensemble output vector \bm{$v$} 
\STATE \bm{$v$} = {$f_{c}f_{m}f_{m-1}...f_{0}(x)$}
\IF {$i$ is the first split point and $i$ $\in$ $sp$}
    \FOR{$k=i$ to $n$} 
    \STATE \bm{$v$} += {$f_{c_{k}}$}{$g_{m}g_{m-1}...g_{k+1}f_{k}f_{k-1}...f_{i}f_{i-1}...f_{1}f_{0}(x)$}
    \ENDFOR 
\ENDIF
\end{algorithmic} 
\end{algorithm}
\par When the split points of the ESD-MBENet-v1 gradually move backwards, the weight-sharing of multi-branch continue to increase, and the number of branches that can be divided decrease, which is very unfriendly to the extraction of image features. Therefore, we propose ESD-MBENet-v2. Compared with ESD-MBENet-v1, ESD-MBENet-v2 can add any number of branches flexibly, and will not be affected by the movement of split points. Suppose, the split point of ESD-MBENet-v2 is set to ``layer2'', we can build four branches as shown in Fig.~\ref{Fig3}. To ensure the diversity of multi-branch output features, although the same backbone is used in four branches, each sub-branch adds a different attention module, that is, the main-branch does not add any additional module, and the sub-branch1 adds SE module, sub-branch2 adds CAM module, and sub-branch3 adds Dropout module. And there is no weight-sharing in the four branches. Assuming that the main-branch is divided into ``conv'', ``main-layer1'', ``main-layer2'', ``main-layer3'' and ``main-layer4'', mathematically expressed as ${f_{0}}$,${f_{1}}$,${f_{2}}$,${f_{3}}$,${f_{4}}$. The sub-branches are denoted as ${l_{1}}$,${l_{2}}$ and ${l_{3}}$. The four branches ${b_{1}}$,${b_{2}}$,${b_{3}}$,${b_{4}}$ divided by ESD-MBENet-v2 are $b_{1} = {f_{4}}{f_{3}}{f_{2}}{f_{1}}{f_{0}}(x),b_{2} = {l_{0}}{f_{2}}{f_{1}}{f_{0}}(x),b_{3} = {l_{1}}{f_{2}}{f_{1}}{f_{0}}(x),b_{4} = {l_{2}}{f_{2}}{f_{1}}{f_{0}}(x)$, which \bm$x$ is a batch of images. The ESD-MBENet-v2
ensemble multi-branch algorithm is shown in Alg.~\ref{alg2}.
\begin{algorithm} 
\caption{ESD-MBENet-v2 multi-branch algorithm} 
\label{alg2} 
\begin{algorithmic}[1] 
\REQUIRE A batch of input images \bm{$x$}. Define main-branch blocks function as a list $[f_{0},f_{1},...,f_{m}]$, split points after the corresponding layer as a list $sp=[0,1,...,n], n \le m-1$, the main-branch fully connected layer {$f_{c}$}, the {$k^{th}$} sub-branch fully connected layer {$f_{c_{k}}$}, the function of the {$k^{th}$} sub-branch {$l_{k}$}, the total branches number $N$.
\ENSURE the ensemble output vector \bm{$v$} \\
\STATE\bm{$v$} = {$f_{c}f_{m}f_{m-1}...f_{1}f_{0}(x)$}\\
\IF {$i$ is the split point and $i$  $\in$ $sp$}
    \FOR{$k=1$ to $N-1$} 
        \STATE{\bm{$v$} += {$f_{c_{k}}$}{$l_{k}f_{i}f_{i-1}...f_{1}f_{0}(x)$}}
    \ENDFOR
\ENDIF
\end{algorithmic} 
\end{algorithm}
\subsection{Self-Distillation}
Using compact multi-branch ensemble network, the accuracy of ESD-MBENet for RS image classification can be significantly improved. To shorten the time of inference and simplify the complexity of ESD-MBENet, we propose to use self-distillation for ESD-MBENet to improve the inference performance of the main-branch. So we can prune all the sub-branches and use only the main-branch for inference.
\par In ESD-MBENet, the self-distillation process includes two parts, final output logits distillation and feature maps distillation. With respect to the output logits distillation, we can get the ensemble output logits from the multiple branches as Eq.~\ref{eq2} and then let it pass through the softmax function as Eq.~\ref{eq3}, which can be as the teacher, and the output logits of the main-branch is as the student. We use KL loss to optimize it. The self-distillation output logits algorithm is shown in Alg.~\ref{alg3}.
\begin{equation}
{v_{t}} = \frac{1}{N}*{\sum_{i=1}^{N}{v_{i}}}
\label{eq2}
\end{equation}
\begin{equation}
p(\hat{y}=y_{i}|\bm{x}) = \frac{e^{v_{i}}}{\sum_{j=1}^{M}{ e^{v_{j}}}}
\label{eq3}
\end{equation}
where $v_{i}$ is the logits of the $i^{th}$ branch, $N$ is the number of all branches, $M$ is the number of all classes.
\begin{algorithm} 
\caption{Self-Distillation output logits algorithm} 
\label{alg3} 
\begin{algorithmic}[1] 
\REQUIRE The total number of branches {$N$}, the {$i^{th}$} branch output logits $\bm{v}_{i}$, main-branch output logits $\bm{v}_{s}$.
\ENSURE the self-distillation loss {$L_{em}^{KL}$} between ensemble output logits and main-branch output logits
\STATE Compute the $\bm{v}_{t}$ using Eq.\ref{eq2} and Eq.\ref{eq3}
\STATE Compute the {$L_{em}^{KL},{\bm{p}_{e}}=\bm{v}_{t},{\bm{p}_{m}}=\bm{v}_{s}$} using Eq.\ref{eq12}
\end{algorithmic} 
\end{algorithm}
\par With respect to the feature maps distillation, the output feature maps (C×H×W) of the multiple branches after ``layer4'' are added along the channel direction to obtain new feature maps (H×W) as Eq.~\ref{eq4}. 
\begin{equation}
    {\bm{g}_{k}^{H*W}} = \sum\limits_{c=1}^{C}{\bm{f}_{k_{c}}^{H*W}}, k=1,2,...,N
    \label{eq4}
\end{equation}
where ${\bm{f}_{k_{c}}^{H*W}}$ is the $k^{th}$ branch feature map in the $c^{th}$ channel, $C$ is the total numble of channels, ${\bm{g}_{k}^{H*W}}$ is the $k^{th}$ branch new feature map. 
\begin{equation}
    {x_{a} = \sum\limits_{i=1}^{H}{\sum\limits_{j=1}^{W}{{g}_{{k}_{ij}}}}/(H*W})
    \label{eq5}
\end{equation}
\begin{equation}
    {x_{s} = \sqrt{\sum\limits_{i=1}^{H}{\sum\limits_{j=1}^{W}{(g_{{k}_{ij}}-x_{a})^{2}}}/(H*W})}
    \label{eq6}
\end{equation}
\begin{equation}
    {F_{k_{ij}}
     = (g_{k_{ij}}-x_{a})/x_{s}}
    \label{eq7}
\end{equation}
\begin{equation}
    {\bm{F}_{k}^{H*W}
     = {F_{k_{ij}}|}_{i=1,2,...,H,j=1,2,...,W}}
    \label{eq8}
\end{equation}
\par Then normalize each of the feature maps (H×W). The normalization process is each pixel value $g_{k_{ij}}$ on the feature map subtracts the mean value $x_{a}$ and then divides the standard deviation $x_{s}$ as Eq.~\ref{eq5}, Eq.~\ref{eq6}, Eq.~\ref{eq7},Eq.~\ref{eq8}. Then we can obtain normalized feature maps $\bm{F}_{k}^{H*W},k=1,2,...,N$. We average the these feature maps to obtain a teacher feature map $\bm{F}_{e}^{H*W}$ as Eq.\ref{eq9}. The student is the main-branch feature map after normalization $\bm{F}_{m}^{H*W}$. The distribution of the feature maps directly affects the output logits. Therefore, if the main-branch feature map (student) can learn the equivalent knowledge to the multi-branch ensemble feature map (teacher), the overall performance of the main-branch will be improved. MSE loss is used to optimize it. Compared with the feature map mutual learning mechanism proposed by previous researchers, we propose a simple feature map learning mechanism, as shown in Alg.~\ref{alg4}.
\begin{equation}
    {\bm{F}_{e}^{H*W}
     = \sum\limits_{k=1}^{N}\bm{F}_{k}^{H*W}/N}
    \label{eq9}
\end{equation}
\begin{algorithm} 
\caption{Self-Distillation feature maps algorithm} 
\label{alg4} 
\begin{algorithmic}[1] 
\REQUIRE Feature maps from the {$k^{th}$} branch {$\bm{f}_{k}^{C*H*W}$}. Define the {$k^{th}$} branch the {$c^{th}$} channel feature map {$\bm{f}_{k_{c}}^{H*W}$}, the total branches {$N$},  the value in row {$i$} and column {$j$} of the feature map {$g_{k_{ij}}$}, after normalizing the {$k^{th}$} branch feature map {$\bm{F}_{k}^{H*W}$},the main-branch feature map after normalizing {$\bm{F}_{m}^{H*W}$} 
\ENSURE the self-distillation loss {$L_{em}^{MSE}$} between ensemble feature map and main-branch feature map
\STATE Compute the new feature map of the $k^{th}$ branch using Eq.~\ref{eq4}.
\STATE Compute the average value of the $k^{th}$ branch new feature map using Eq.~\ref{eq5}.
\STATE Compute the standard deviation value of the $k^{th}$ branch new feature map using Eq.~\ref{eq6}.
\STATE Compute the feature map of the $k^{th}$ branch after normalization using Eq.~\ref{eq7} and Eq.~\ref{eq8}.
\STATE Compute the ensemble feature map of the all branches after normalization using Eq.~\ref{eq9}.
\STATE Compute the {$L_{em}^{MSE}$}, ${\bm{F}_{e}}(\bm{x})={\bm{F}_{e}^{H*W}},{\bm{F}_{m}}(\bm{x})={\bm{F}_{m}^{H*W}}$ using Eq.~\ref{eq13}.
\end{algorithmic} 
\end{algorithm}
\subsection{Backward Propagation of ESD-MBENet}
The ESD-MBENet we proposed is optimized by continuously reducing the total loss objective function as Eq.~\ref{eq10}.
We use cross-entropy loss as Eq.~\ref{eq11}, Kullback Leibler divergence (KL) loss as Eq.~\ref{eq12} and Mean Square Error (MSE) loss as Eq.~\ref{eq13} to make ESD-MBENet converge quickly.
\begin{equation}
    L_{total} = \sum\limits_{i=1}^{N}(\alpha_{i}*L_{ce_{i}}) + \beta*L_{em}^{KL}+\lambda*L_{em}^{MSE}
    \label{eq10}
\end{equation}
\begin{equation}
    L_{ce_{i}} = -\sum\limits_{j=1}^{M}(y_{ij}\text{log} (p_{ij}))
    \label{eq11}
\end{equation}
\begin{equation}
\begin{split}
    \begin{aligned}
        L_{em}^{KL} &= L^{KL}({\bm{p}_{e}||\bm{p}_{m}}) = -\frac{1}{n}\sum\limits_{i=1}^{n}{\bm{p}(\bm{x}_{ei})\text{log}\frac{\bm{p}(\bm{x}_{mi})}{\bm{p}(\bm{x}_{ei})}}\\ 
        &= {H(\bm{p_{e},p_{m}})-H(\bm{p_{e}})}
    \end{aligned}
\end{split}
\label{eq12}
\end{equation}

\begin{small}
\begin{equation}
 \begin{split}
   \emph L_{em}^{MSE} &= \frac{1}{T} \sum_{t=1}^{T} \emph L^{MSE} \big{(} \bm{F_{e}}(\bm{x}_t) \| \bm{F_m}(\bm{x}_t) \big{)}  \\
                  &= \frac{1}{T} \sum_{t=1}^{T} \sum_{i=1}^{H}\sum_{j=1}^{W} {\big(} (f_{e}(\bm{x}_t))_{ij}-(f_m(\bm{x}_t))_{ij} \big{)}^2
   \end{split}
   \label{eq13}
\end{equation}
\end{small}

where {$L_{ce_{i}}$} represents the cross-entropy loss function, and the cross-entropy loss is obtained from each branch respectively, {$L_{em}^{KL}$} represents the KL loss between ensemble output logits $\bm{p}_{e}$ and the main-branch output logits $\bm{p}_{m}$, $L_{em}^{MSE}$ is the MSE loss between the ensemble feature map $\bm{F_{e}}(\bm{x})$ and the main-branch feature map $\bm{F_m}(\bm{x})$. {$\alpha$}, {$\beta$} and {$\lambda$} are the weight coefficients of each loss function. Our proposed ESD-MBENet network uses only two loss functions in the distillation process of output logits and feature maps, regardless of the number of sub-branches. This greatly simplifies the process of tuning and optimization.
\begin{table*}
\begin{center}
\caption{The detailed structure of baseline models~\cite{VGG,ResNet,densenet}. The layer name corresponds to the layers in Fig.~\ref{Fig2} and Fig.~\ref{Fig3}. VGG16, ResNet50 and DenseNet121 are used for RS image classification. ResNet18/34 and ResNet20/32/44/56 network are used to classify the natural scene image CIFAR/tiny-ImageNet dataset. For VGG16, FC layer channels number is modified to 512. In addition, since CIFAR10/100 and tiny-ImageNet image size are 32×32 and 64×64, we changed the ``conv1'' of ResNet18/34 to 3×3, and set the stride to 1. The number of BasicBlock channels used in ResNet18/34 is different from that used in ResNet20/32/44/56.}
\label{tab1}
\renewcommand\arraystretch{1.5}  
\begin{tabular}{|c|c|c|c|c|c|}
  \cline{1-6}
   \multirow{2}{*}{Layer Name} & 
    \multicolumn{5}{c|}{Network}
  \\ \cline{2-6}
  {} & VGG16 & ResNet50 & DenseNet121 & ResNet18/34 & ResNet20/32/44/56
  \\ \cline{1-6}
  Conv1 & conv1-x × 2 & \multicolumn{2}{c|}{7 × 7 conv, stride 2} & \multicolumn{2}{c|}{3 × 3 conv, stride 1}
  \\ \cline{1-6}
  Pool1 & 2×2 max pool, stride 2 & \multicolumn{2}{c|}{3 × 3 max pool, stride 2} & \multicolumn{2}{c|}{-}
  \\ \cline{1-6}
  Layer1 & conv2-x × 2 & Bottleneck × 3 & DenseBlock × 6 & BasicBlock × 2/3 & BasicBlock × 3/5/7/9
  \\ \cline{1-6}
  Layer2 & conv3-x × 3 & Bottleneck × 4 & Transition,DenseBlock × 12 & BasicBlock × 2/4 & BasicBlock × 3/5/7/9
  \\ \cline{1-6}
  Layer3 & conv4-x × 3 & Bottleneck × 6 & Transition,DenseBlock × 24 & BasicBlock × 2/6 & BasicBlock × 3/5/7/9
  \\ \cline{1-6}
  Layer4 & conv5-x × 3 & Bottleneck × 3 & Transition,DenseBlock × 16 & BasicBlock × 2/3 & -
  \\ \cline{1-6}
  Pool2 & \multicolumn{5}{c|}{average pool}
  \\ \cline{1-6}
  FC  & FC1 × 2, 512 × num\_cls & 2048 × num\_cls & 1024 × num\_cls & 512 × num\_cls & 64 × num\_cls
  \\ \cline{1-6}
\end{tabular}
\end{center}
\end{table*}
\section{Experiments}
\subsection{Datasets}
We use three RS datasets (AID, NWPU-RESISC45, UC-Merced) to verify the effectiveness of ESD-MBENet in RS images classification to compare with other methods conveniently. The AID dataset has 10,000 RS images, including 30 categories, and the image size is 600×600. There are 220$\sim$420 images per category. This dataset was released by Huazhong University of Science and Technology and Wuhan University in 2017. The NWPU-RESISC45 dataset has 31,500 images, including 45 categories, each category has 700 images, and the image size is 256×256. The dataset was created by Northwestern Polytechnical University. The UC-Merced dataset has only 2100 images, including 21 categories. The image size is 256×256, and each category has 100 images. This dataset is extracted from USGS National Map Urban Area Imagery. In addition, to verify the generalization of ESD-MBENet, we also select the natural scene image classification datasets (CIFAR, tiny-ImageNet) to do related experiments. 
\subsection{Implementation Details}
We select three backbones, namely VGG16, ResNet50, DenseNet121 on three datasets (AID,NWPU-RESISC45,UC-Merced) to expriment. The detail structure of the backbone is shown in Tab.~\ref{tab1}. Since most of the methods proposed by many researchers used VGG16 as the backbone, to facilitate comparison with them, we select VGG16 as the backbone. ResNet50 was proposed by He Kaiming~\cite{ResNet}. It has superior performance than VGG16 and is widely used. Therefore, we also select ResNet50 as one of our backbones. For deeper networks, we select DenseNet121, which can compare with the SOTA results of KFBNet~\cite{KFB}. The optimizer used in the experiment is the Stochastic Gradient Descent (SGD) with momentum, and the momentum parameter is set to 0.9. The image is resized to 256×256 during the training of the AID dataset, and resized to 288×288 and croped to 256×256 during the test. The models are trained for 100 epoches in each experiment on AID dataset, and the learning rate drops 10 times at the 40th, 70th and 90th epoch. The training images of the NWPU-RESISC45 and UC-Merced are resized to 224×224, and the test images are resized to 256×256 and croped to 224×224. The models are trained for 120 epoches in each experiment on NWPU-RESISC45 dataset, and the learning rate drops 10 times at the 70th, 90th, and 110th epoch. ImageNet pretrained parameters are loaded in each layer when training. The code is implemented using the Pytorch framework. The equipment used in the experiment is NVIDIA GTX 1080ti.

\begin{table*}
\begin{center}
\caption{Comparison of classification results (\%) on AID, NWPU-RESISC45 and UC-Merced. ``tr'' denotes training rate.}
\label{tab2}
\renewcommand\arraystretch{1.5}
\begin{tabular}{|c|c|c|c|c|c|c|}
  \cline{1-7}
    \multirow{2}{*}{Networks} & \multirow{2}{*}{Backbone} & 
    \multicolumn{2}{c|}{AID} &  \multicolumn{2}{c|}{NWPU-RESISC45} & UC-Merced
  \\ \cline{3-7}
  {} & {} & tr = 20\% & tr = 50\% & tr = 10\% & tr = 20\% & tr = 80\%
  \\ \cline{1-7}
  DCNN~\cite{DCNN} & VGG16 & 90.82$\pm$0.16 & 96.89$\pm$0.10 & 89.22$\pm$0.50 & 91.89$\pm$0.22 & 98.93$\pm$0.10
  \\ \cline{1-7}
  MSCP~\cite{MSCP} & VGG16 & 92.21$\pm$0.17 & 96.56$\pm$0.18 & 88.07$\pm$0.18 & 90.81$\pm$0.13 & 98.40$\pm$0.34
  \\ \cline{1-7}
  RTN~\cite{att_RS1} & VGG16 & 92.44 & - & 89.90 & 92.71 & 98.96
  \\ \cline{1-7}
  MG-CAP~\cite{MGCAP} & VGG16 & 93.34$\pm$0.18 & 96.12$\pm$0.12 &  $ \textbf{90.83}\pm\textbf{0.12}$ & 92.95$\pm$0.13 & 99.0$\pm$0.10
  \\ \cline{1-7}
  SCCov~\cite{SCCov} & VGG16 & 93.12$\pm$0.25 & 96.10$\pm$0.16 & 89.30$\pm$0.35 & 92.10$\pm$0.25 & 99.05$\pm$0.25
  \\ \cline{1-7}
  Hydra\cite{Hydra} & DenseNet121 & - & - & 92.44$\pm$0.34 & 94.51$\pm$0.21 & -
  \\ \cline{1-7}
  KFBNet~\cite{KFB} & VGG16 & 94.27$\pm$0.02 & 97.19$\pm$0.07 & 90.27$\pm$0.02 & 92.54$\pm$0.03 & $\textbf{99.76}\pm\textbf{0.24}$
  \\ \cline{1-7}
  KFBNet~\cite{KFB} & DenseNet121 & 95.50$\pm$0.27 & 97.40$\pm$0.10 & 93.08$\pm$0.14  & 95.11$\pm$0.10 & $\textbf{99.88}\pm\textbf{0.12}$
  \\ \cline{1-7}
  ESD-MBENet-v1 & VGG16 & 94.10$\pm$0.13 & 97.15$\pm$0.21 & $\textbf{90.29}\pm\textbf{0.11}$ & $\textbf{93.48}\pm\textbf{0.06}$ & $\textbf{99.81}\pm\textbf{0.10}$
  \\ \cline{1-7}
  ESD-MBENet-v2 & VGG16 & 94.12$\pm$0.24 & $\textbf{97.3}\pm\textbf{0.08}$ & 90.25$\pm$0.21 & $\textbf{93.42}\pm\textbf{0.15}$ & 
  $\textbf{99.86}\pm\textbf{0.12}$
  \\ \cline{1-7}
  ESD-MBENet-v1 & ResNet50 & $\textbf{96.0}\pm\textbf{0.15}$ & $\textbf{98.54}\pm\textbf{0.17}$ & 92.5$\pm$0.22 & $\textbf{95.58}\pm\textbf{0.08}$ & 
  $\textbf{99.81}\pm\textbf{0.10}$
  \\ \cline{1-7}
  ESD-MBENet-v2 & ResNet50 &
  $\textbf{95.81}\pm\textbf{0.24}$ & $\textbf{98.66}\pm\textbf{0.2}$ & 
  93.03$\pm$0.11 
  & $\textbf{95.24}\pm\textbf{0.23}$ & 
  $\textbf{99.86}\pm\textbf{0.12}$
  \\ \cline{1-7}
  ESD-MBENet-v1 & DenseNet121 & $\textbf{96.2}\pm\textbf{0.15}$ & $\textbf{98.85}\pm\textbf{0.13}$ & $\textbf{93.24}\pm\textbf{0.15}$
  & $\textbf{95.5}\pm\textbf{0.09}$ &
  $\textbf{99.86}\pm\textbf{0.12}$
  \\ \cline{1-7}
  ESD-MBENet-v2 & DenseNet121 & $\textbf{96.39}\pm\textbf{0.21}$ & $\textbf{98.4}\pm\textbf{0.23}$ & 93.05$\pm$0.18 & $\textbf{95.36}\pm\textbf{0.14}$ & $\textbf{99.81}\pm\textbf{0.10}$
  \\ \cline{1-7}
\end{tabular}
\end{center}
\end{table*}

\subsection{Results}
The experimental results are shown in the Tab.~\ref{tab2}. We compare the ESD-MBENet with the previously proposed excellent algorithm when using the same backbone, the same dataset. To reduce the experimental error, we did each experiment five times, and reported the results as the mean and standard deviation of the five experiments.
\subsubsection{AID}
To make a fair comparison with the previous methods, the setting of ESD-MBENet in the AID dataset is the same as them. That is, 20\% of the data is randomly selected as the training set, 80\% as the test set or 50\% as the training set and 50\% as the test set. If the backbone is VGG16, the results of ESD-MBENet-v1 and ESD-MBENet-v2 are 94.10\% and 94.12\% on 20\% training data. Although it did not exceed SOTA, it was very close to it. On 50\% training data the results are 97.15\% and 97.3\%. When using ResNet50 as the backbone and 20\% and 50\% of the data as the training set, ESD-MBENet-v1 can reach 96.0\% and 98.54\% accuracy, ESD-MBENet-v2 can reach 95.81\% and 98.66\%; when DenseNet121 is selected as the backbone, ESD-MBENet-v1 can achieve accuracy rates of 96.2\% and 98.85\%, and ESD-MBENet-v2 can achieve accuracy rates of 96.39\% and 98.4\%. Compared with KFBNet, we did not add additional elements to the network in the inference stage, that is, only the main-branch is used for inference, but the results exceeded about 1\% in DenseNet121.
\subsubsection{NWPU-RESISC45}
In the experiment, we randomly select 20\% for training, 80\% for testing or 10\% for training and 90\% for testing in all datasets. If VGG16 is the backbone of the ESD-MBENet, 10\% and 20\% data for training, the accuracy of the ESD-MBENet-v1 is 90.29\% and 93.48\%, the accuracy of the ESD-MBENet-v2 is 90.25\% and 93.42\%. If ResNet50 is the backbone of the ESD-MBENet, we can achieve the accuracy of 92.5\% and 95.58\% in ESD-MBENet-v1 and the accuracy of 93.03\% and 95.24\% in ESD-MBENet-v2. If the backbone is DenseNet121, ESD-MBENet-v1 can reach 93.24\% and 95.5\%, ESD-MBENet-v2 can reach 93.05\% and 95.36\%. In most cases, ESD-MBENet reaches the SOTA. In a few cases, although ESD-MBENet did not reach the SOTA, it is very close to the previous SOTA results. And compared with previous methods ESD-MBENet is simpler in inference.
\subsubsection{UC-Merced}
The accuracy of proposed methods previously in the UC-Merced dataset has reached the limit. The ESD-MBENet we proposed is the same. During the experiment, even if VGG16 is used as the backbbone, the accuracy can reach 100\% sometimes. There are a total of 420 test images. This result means that at most one image can be predicted incorrectly or all predictions are accurate in each experiment. The average accuracy of multiple test can reach 99.81\% or 99.86\%.
\subsubsection{CIFAR-10/100 and tiny-ImageNet}
In order to verify the generalization performance of the ESD-MBENet, we tried to experiment on the common scene classification dataset CIFAR-10/100 and tiny-ImageNet. During the experiment, we select ResNet20/32/44/56 and ResNet18/34 as backbones, and the detailed structure is in Tab.~\ref{tab1}. ResNet20/32/44/56 represents a shallower network, and ResNet18/34 represents a deeper network. The experiment process is the same as that used in RS images. 
\par As shown in the Tab.\ref{tab3}, the results of our proposed ESD-MBENet on the CIFAR-10/100 are higher than baseline network in all backbones, especially the CIFAR-100, the experimental results are higher more than 3\%. Compared the previous methods~\cite{cifar-sota,cifar-sota1}, our result is also higher more than 1\% and reaches the SOTA. Tab.~\ref{tab4} shows the ESD-MBENet experimental results of ResNet18 and ResNet34 on the CIFAR-100 and tiny-ImageNet. Our results on the tiny-ImageNet dataset exceed the baseline by 3\%$\sim$5\%.
\begin{table}
\caption{The top-1 error of the baseline model of ResNet20/32/44/56 and ESD-MBENet-v1/v2 on CIFAR-100/10.``v1'' denotes ESD-MBENet-v1, ``v2'' denotes ESD-MBENet-v2.}
  \label{tab3}
  \begin{center}
  \renewcommand\arraystretch{1.5}
  \begin{tabular}{|c| c |c |c |c| c| c|}
  \cline{1-7}
  \multirow{2}{*}{Methods}    &\multicolumn{3}{c|}{CIFAR-100}        &\multicolumn{3}{c|}{CIFAR-10}   \\
  \cline{2-7}
         &Baseline   &v1  &v2
         &Baseline   &v1  &v2      \\
  \cline{1-7}
  ResNet20              
  &32.66    &28.54  &28.79  
  &8.37     &6.69   &6.77                              \\
  \cline{1-7}

  ResNet32              
  &30.73    &26.46  &26.69  
  &7.35     &6.03   &6.07                               \\
  \cline{1-7}

  ResNet44              
  &29.43    &25.74  &26.08  
  &6.78     &5.83   &5.88                              \\

  \cline{1-7}

  ResNet56              
  &28.91    &25.63  &25.40  
  &6.13     &5.49   &5.48                              \\
  \cline{1-7}
  \end{tabular}
  \end{center}
\end{table}
\begin{table}
\begin{center}
\caption{The top-1 error of the baseline model of ResNet18/34 and ESD-MBENet-v1/v2 on CIFAR-100 and tiny-ImageNet datasets.}
\label{tab4}
\renewcommand\arraystretch{1.5}
\begin{tabular}{|c |c |c|}
  \cline{1-3}
  Methods(backbone) & CIFAR-100 & tiny-ImageNet   \\
  \cline{1-3}
  ResNet18  & 23.71 & 30.91 \\
  \cline{1-3}
  ESD-MBENet-v1(ResNet18) & 20.49  & 26.81 \\
  \cline{1-3}
  ESD-MBENet-v2(ResNet18) & 20.51  & 25.86 \\
  \cline{1-3}
  ResNet34  & 22.16 & 23.83 \\
  \cline{1-3}
  ESD-MBENet-v1(ResNet34) & 20.32  & 21.74 \\
  \cline{1-3}
  ESD-MBENet-v2(ResNet34) & 20.09  & 21.12 \\
  \cline{1-3}
  \end{tabular}
\end{center}
\end{table}
\subsection{Confusion Matrix}
\begin{figure}[ht]
\centering
\includegraphics[width=1.0\linewidth]{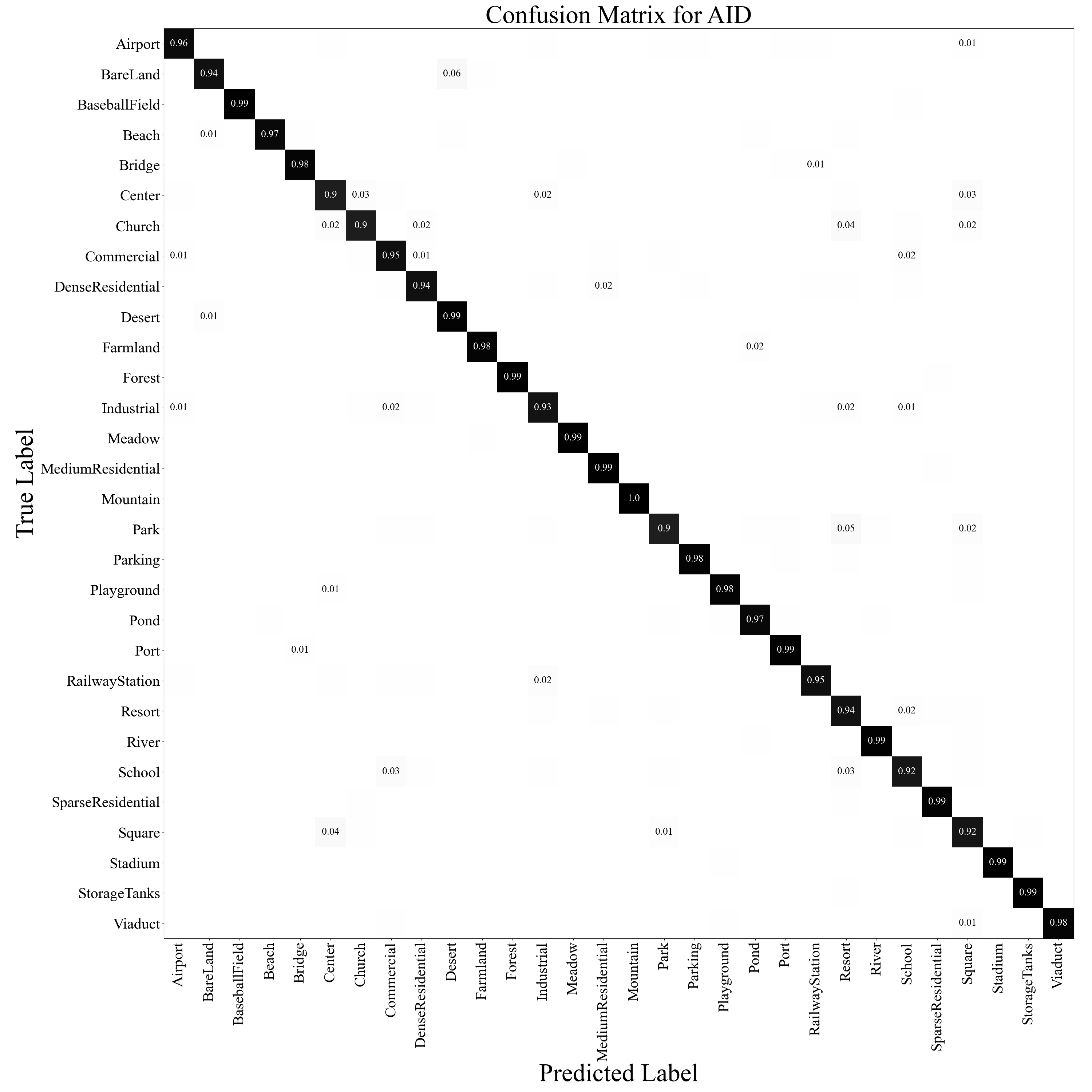}
\caption{ESD-MBENet-v1 confusion matrix of DenseNet121 on 20\% AID training data.}
\label{Fig5}
\end{figure}

\begin{figure}[ht]
\centering
\includegraphics[width=1.0\linewidth]{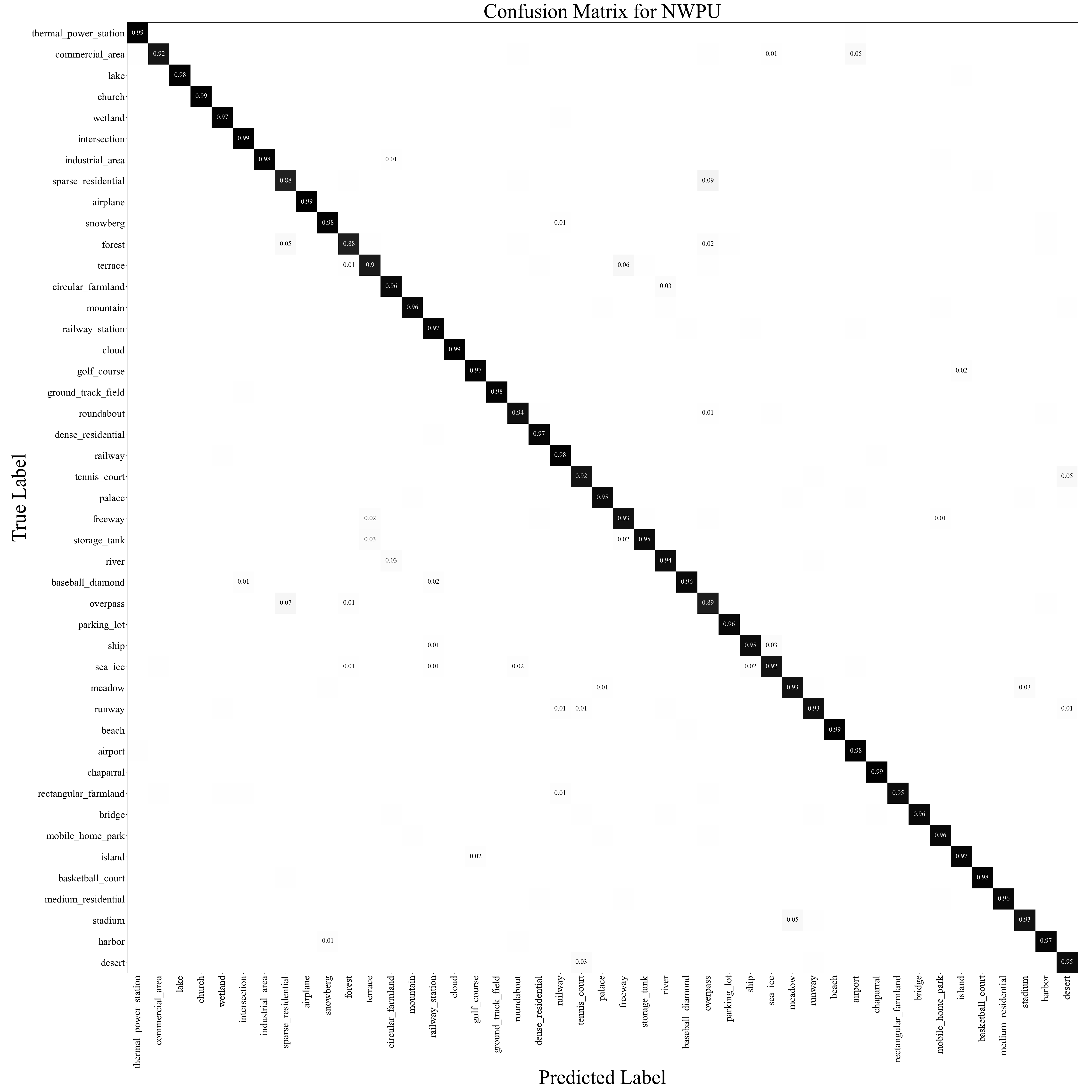}
\caption{ESD-MBENet-v1 confusion matrix of ResNet50 on 20\% NWPU-RESISC45 training data.}
\label{Fig6}
\end{figure}

In RS image scene classification, confusion matrix is usually used as an evaluation criterion to judge the effect of the proposed algorithm. The confusion matrix can be used to represent the difference between the predicted label and the true label. From the confusion matrix, you can see how much data in the predicted label is correctly predicted and how many predicted labels are predicted incorrectly. And you can also see what the result of the prediction error should be, but what label is actually predicted. The confusion matrix is very intuitive and clear, which is very useful for data analysis. In this paper, a confusion matrix is used to check the effectiveness of ESD-MBENet. The Fig.~\ref{Fig5} shows the confusion matrix of 20\% AID training data in DenseNet121 network. It can be seen from the confusion matrix that ESD-MBENet-v1 has a less than 4\% prediction error rate for almost all classes, and the prediction accuracy rate of some classes even reaches 99\% and 100\%. Fig.~\ref{Fig6} shows the confusion matrix made by ESD-MBENet-v1(ResNet50) prediction results on randomly selected 20\% NWPU-RESISC45 training dataset. The abscissa of the confusion matrix represents the predicted label, and the ordinate represents the true label.
\begin{table}
\begin{center}
\caption{Comparison of experimental results between baseline network and ESD-MBENet on the AID and NWPU-RESISC45 datasets. ``v1-OD'' and ``v2-OD'' denote ESD-MBENet only self-distill the output logits and not self-distill the feature maps.}
\label{tab5}
\renewcommand\arraystretch{1.5}
\scalebox{0.8}{
\begin{tabular}{|c|c|c|c|c|}
  \cline{1-5}
  \multirow{2}{*}{Networks} & \multicolumn{2}{c|}{AID} & \multicolumn{2}{c|}{NWPU-RESISC45}
  \\ \cline{2-5}
  {} & tr = 20\% & tr = 50\% & tr = 10\% & tr = 20\%
  \\ \cline{1-5}
 VGG16 & 93.21$\pm$0.24 & 96.9$\pm$0.16 & 88.16$\pm$0.13 & 92.69$\pm$0.09    
  \\ \cline{1-5}
 VGG16-v1-OD & 93.82$\pm$0.15 & 97.08$\pm$0.15 & 89.98$\pm$0.21 & 93.15$\pm$0.17
 \\ \cline{1-5}
 VGG16-v1 & 94.10$\pm$0.13 & 97.15$\pm$0.21 & 90.29$\pm$0.11 & 93.48$\pm$0.06
 \\ \cline{1-5}
 VGG16-v2-OD & 94.03$\pm$0.22 & 96.98$\pm$0.23 & 89.95$\pm$0.25 & 93.12$\pm$0.2
  \\ \cline{1-5}
 VGG16-v2 & 94.12$\pm$0.24 & 97.3$\pm$0.08 & 90.25$\pm$0.21 & 93.42$\pm$0.15
  \\ \cline{1-5}
 ResNet50 & 95.4$\pm$0.05 & 97.82$\pm$0.12 & 91.6$\pm$0.09 & 94.85$\pm$0.14
  \\ \cline{1-5}
 ResNet50-v1-OD & 95.75$\pm$0.23 & 98.44$\pm$0.2 & 92.48$\pm$0.24 & 95.2$\pm$0.12
 \\ \cline{1-5}
 ResNet50-v1 & 96.0$\pm$0.15 & 98.54$\pm$0.17 & 92.5$\pm$0.22 & 95.58$\pm$0.08
 \\ \cline{1-5}
 ResNet50-v2-OD & 95.80$\pm$0.22 & 98.22$\pm$0.13 & 92.91$\pm$0.19 & 95.15$\pm$0.21
  \\ \cline{1-5}
 ResNet50-v2 & 95.81$\pm$0.24 & 98.66$\pm$0.2 & 93.03$\pm$0.11 & 95.24$\pm$0.23
 \\ \cline{1-5}
 DenseNet121 & 95.46$\pm$0.13 & 98.2$\pm$0.07 & 91.77$\pm$0.22 & 94.34$\pm$0.15
  \\ \cline{1-5}
 DenseNet121-v1-OD & 95.96$\pm$0.19 & 98.8$\pm$0.24 & 92.8$\pm$0.24 & 95.42$\pm$0.2
 \\ \cline{1-5}
 DenseNet121-v1 & 96.2$\pm$0.15 & 98.85$\pm$0.13 & 93.24$\pm$0.15 & 95.5$\pm$0.09
 \\ \cline{1-5}
 DenseNet121-v2-OD & 96.30$\pm$0.18 & 98.38$\pm$0.06 & 92.92$\pm$0.22 & 95.23$\pm$0.16
  \\ \cline{1-5}
 DenseNet121-v2 & 96.39$\pm$0.21 & 98.4$\pm$0.23 & 93.05$\pm$0.18 & 95.36$\pm$0.14
 \\ \cline{1-5}
\end{tabular}}
\end{center}
\end{table}
\begin{table}
\begin{center}
\caption{Compared the multi-branch ensemble output with main-branch output on 20\% training data of the NWPU-RESISC45 in inference, which is used DenseNet121 as the backbone. ``ESD-MBENet-v1-E'' and ``ESD-MBENet-v2-E'' denote the multi-branch ensemble output of the ESD-MBENet.}
\label{tab6}
\renewcommand\arraystretch{1.5}

  \begin{tabular}{|c |c |c|c|}
  \cline{1-4}
  Methods & Accuracy & Parameters & FLOPs  \\
  \cline{1-4}
  ESD-MBENet-v1-E  & 95.52$\pm$0.12  & 14.08M & 7.96G \\
  \cline{1-4}
  ESD-MBENet-v1  & 95.5$\pm$0.09 & 7.98M & 2.87G \\
  \cline{1-4}
  ESD-MBENet-v2-E & 95.44$\pm$0.17 & 23.88M & 5.16G \\
  \cline{1-4}
  ESD-MBENet-v2 & 95.36$\pm$0.14 & 7.98M & 2.87G \\
  \cline{1-4}
  \end{tabular}
\end{center}
\end{table}
\begin{figure*}
  \centering
  \includegraphics[width=1.0\linewidth]{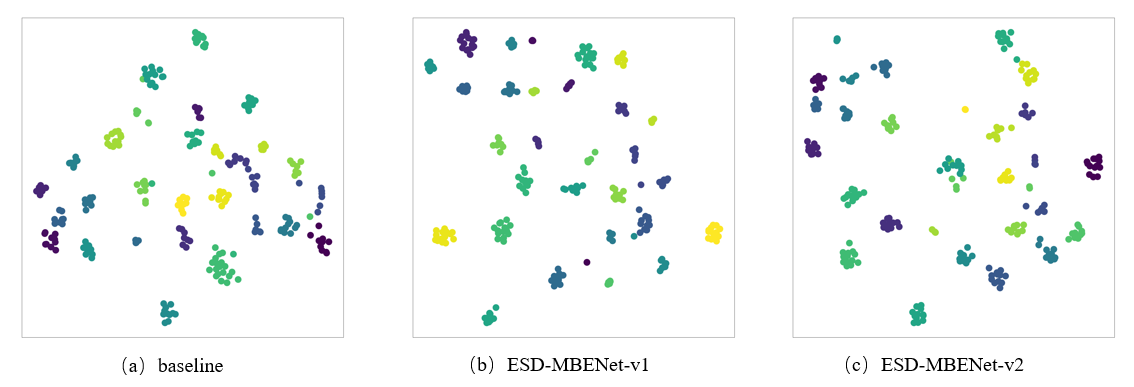}
  \caption{T-SNE of the DenseNet121 baseline network, ESD-MBENet-v1 and ESD-MBENet-v2. There are more inter-class differences in ESD-MBENet-v1 and ESD-MBENet-v2 compared with the baseline network. Different colors indicate different categories, and the same categories are gathered into a small pile. The larger the distance between the small piles, the larger the gap in different classes and the better the classification effect.}
\label{Fig7}
\end{figure*}
\begin{figure}[ht]
\centering
\includegraphics[width=1.0\linewidth]{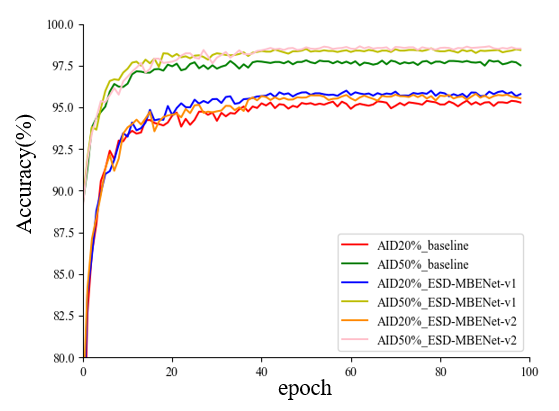}
\caption{The training curves of AID on ResNet50. The accuracy changes of the training process of baseline, ESD-MBENet-v1 and ESD-MBENet-v2 were compared on the 20\% and 50\% AID training datasets respectively.  ``AID20\%'' means we select 20\% data for training.}
\label{Fig8}
\end{figure}
\begin{figure*}
  \centering
  \includegraphics[width=1.0\linewidth]{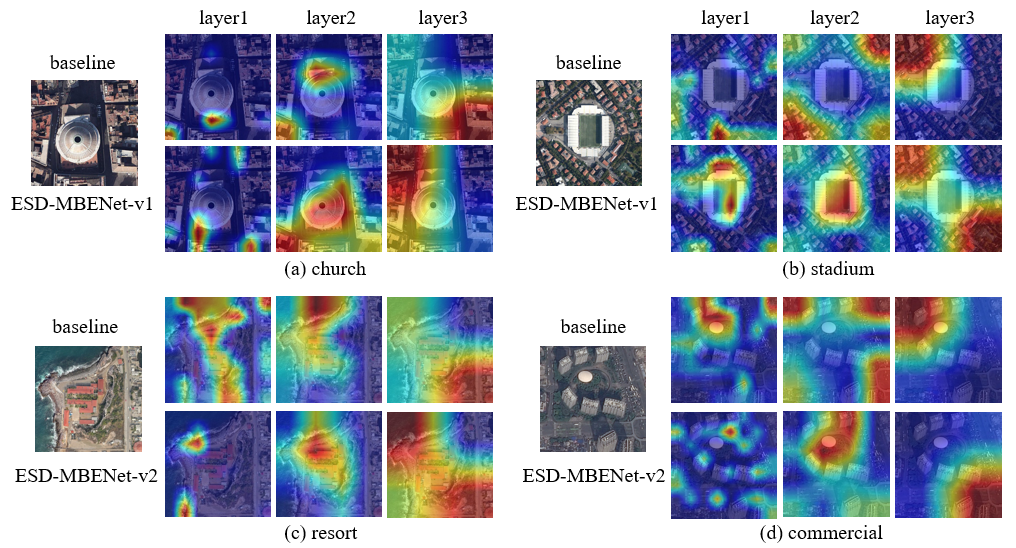}
  \caption{The Grad-CAM comparison of ResNet50 baseline network and ESD-MBENet on the 20\% AID training dataset. We randomly select four images from the AID dataset as representatives, and show the focus of different parts of the baseline and ESD-MBENet, such as ``layer1'', ``layer2'' and ``layer3''  during RS image classification. The warmer the color, the higher the degree of attention.}
\label{Fig9}
\end{figure*}
\subsection{Ablation Study}
To more effectively verify the effects of multi-branch ensemble and self-distillation of our proposed ESD-MBENet algorithm, and the robustness of the network to add different attention modules to sub-branch, we did the following ablation experiments.
\subsubsection{Comparison between ESD-MBENet and baseline}
To solve the interference from different characteristics of different geographical elements in RS images, ESD-MBENet has introduced the method of multi-branch ensemble network in the training process, allowing the network to explore image information from multiple perspectives and get more attention to the classification information of the image. In the inference stage, we introduce self-distillation on the output logits and feature maps to reduce the complexity of the model, there is no difference in inference speed compared with the baseline network. 
\par We use VGG16, ResNet50 and DenseNet121 as the backbones respectively, and do the following comparative experiments on AID and NWPU-RESISC45 datasets. It can be seen from the Tab.~\ref{tab5} that ESD-MBENet-v1 and ESD-MBENet-v2 both have more than 1\% improvement of compared with the baseline network. This can also verify that ESD-MBENet has indeed learned more image information in feature extraction through multi-branch feature ensemble and self-distillation, which is more helpful for RS image classification.
\subsubsection{Self-distillation in ESD-MBENet feature maps}
Distillation technology is essentially a process in which a student network continuously learns and imitates a teacher network to achieve student knowledge enhancement. In this experiment, we use the ESD-MBENet itself as a teacher, so it can be called self-distillation. We mainly use self-distillation in output logits and intermediate feature maps. For students' learning, if the teacher directly tells the students the standard answer every time, let the students explore the learning process by themselves, in most cases the students will learn very well. However, if the teacher also provides guidance and advice in the middle learning process, this may be more helpful to the students' learning. Therefore, we propose to use self-distillation technology in feature maps. In order to effectively compare the effectiveness of the idea, we also did relevant comparative experiments on the AID and NWPU-RESISC45 datasets on the VGG16, ResNet50 and  DenseNet121 networks respectively. As shown in Tab.~\ref{tab5}, students who not only self-distill the output logits,but also self-distill the feature maps learn better than only self-distill the output logits.
\subsubsection{ESD-MBENet outputs of ensemble multi-branch and main-branch}
Distillation is a process of mutual assistance. Therefore, when students learn well, the teacher may also be inspired, although sometimes the effect of this inspiration is small. Therefore, during the experiment, we also compare the results of the multi-branch ensemble with the output of the main-branch, as well as the parameters and FLOPs in inference. As shown in the Tab.~\ref{tab6}, it can be seen that the ensemble teacher outputs does indeed learn a little more knowledge than the main-branch student network, but it consumes more parameters and FLOPs than main-branch in inference. This is not friendly for practical applications. Therefore, considering all factors, we choose to use only the main-branch for inference.
\subsubsection{Comparison of ESD-MBENet-v1 sub-branch using different attention modules}
To verify that our proposed ESD-MBENet is still effective even when the sub-branch uses different attention modules, this is mainly for ESD-MBENet-v1. In the experiment comparison process, we mainly add the SE or CAM or Dropout module in the sub-branch. The experimental results are shown in Tab.~\ref{tab7}. The multi-branch network we constructed is robust to the addition of different attention modules.
\begin{table}
\begin{center}
\caption{Comparison of results of ESD-MBENet-v1 sub-branch using different attention modules, with ResNet50 as the backbone. ``ESD-MBENet-v1-SE'' means that the attention module added to the sub-branch is SE. The training rate of the AID and NWPU-RESISC45 is 20\%.}
\label{tab7}
\renewcommand\arraystretch{1.5}
  \begin{tabular}{|c |c |c|}
  \cline{1-3}
  Methods & NWPU-RESISC45 & AID \\
  \cline{1-3}
  ESD-MBENet-v1-SE  & 95.42$\pm$0.14 & 95.78$\pm$0.21  \\
  \cline{1-3}
  ESD-MBENet-v1-CAM  & 95.38$\pm$0.2 & 95.87$\pm$0.17  \\
  \cline{1-3}
  ESD-MBENet-v1-Dropout & 95.58$\pm$0.08 & 96.0$\pm$0.15 \\
  \cline{1-3}
\end{tabular}
\end{center}
\end{table}
\subsection{Visualization and Analysis}
\subsubsection{Training Curves}
To compare the convergence of ESD-MBENet with the baseline network more intuitively, we plot the curves of the experimental results. As shown in the Fig.~\ref{Fig8}, we use ResNet50 as the backbone to train 20\% or 50\% data of the AID dataset. It can be seen from the curves that the overall performance of the ESD-MBENet is better than the baseline in both ESD-MBENet-v1 and ESD-MBENet-v2. And the difference between the results of ESD-MBENet-v1 and ESD-MBENet-v2 is very small. We set a total of 100 epoches. The accuracy of baseline and ESD-MBENet networks are gradually flattening out around the 50th epoch. Compared with baseline post-training, ESD-MBENet has less fluctuation and is more stable.
\subsubsection{T-SNE}
T-SNE technology can map data in high-dimensional space to low-dimensional space. It can clearly see the difference between different algorithms for RS image classification. Therefore, we use T-SNE technology to show a 2-dimensional mapping representation of the final output results. As shown in Fig.~\ref{Fig7}, we compare the baseline, ESD-MBENet-v1 and ESD-MBENet-v2 networks, which backbone is DenseNet121. Compared with the baseline, ESD-MBENet-v1 and ESD-MBENet-v2 can obtain larger inter-class differences for RS image classification, which is very useful for accurate classification. The more similar the category, the larger the gap between the categories is needed to achieve better classification, and the network will not be confused due to the large gap between the categories. Therefore, ESD-MBENet achieves a better classification effect than baseline.
\subsubsection{Grad CAM}
Grad-CAM is a relatively popular visualization method, which can make it easier for us to understand how convolutional neural networks learn for a given task, such as image classification or image segmentation. We also visualized the Grad-CAM experimental effect of ESD-MBENet. And compared with the baseline network. In the experiment, the baseline, ESD-MBENet-v1 and ESD-MBENet-v2 models trained on the ResNet50 network using the 20\% AID dataset were used to draw Grad-CAM on four randomly selected images. To compare the learning effect of the network at different stages more clearly, we have shown the Grad-CAM of the different depths of the network, such as ``layer1'', ``layer2'' and ``layer3', which can also represent the learning focus of the network in the shallow stage and the deep stage. It can be seen from the Fig.~\ref{Fig9} that the ESD-MBENet network pays more attention to the objects to be classified at the ``layer2'' than the baseline network. At the ``layer3'', the focus of the ESD-MBENet is more than that of the baseline, which means ESD-MBENet can extract more information of the images and then transfer it to the deeper network. This is more conducive to network learning.
\section{Conclusion}
In this paper, we design ESD-MBENet-v1 and ESD-MBENet-v2 to construct compact multi-branch ensemble network to solve the interference from different characteristics of different geographical elements in RS images. ESD-MBENet-v1 uses as few modules as possible to build as many branches as possible, but as the slpit points move backwards, the number of branches built decreases. Therefore, we propose ESD-MBENet-v2, which can build multiple branches flexibly. ESD-MBENet-v2 achieves the greatest possible weight-sharing. Due to the multi-branch construction, although the network performance has been greatly improved, in the inference stage, the model is too complex to reduce the inference efficiency and speed. So we propose self-distillation, distilling the logits and the intermediate feature maps, to make the main-branch network reach the performance of the whole model. In this way, only the main-branch is used for inference. Through experimental verification, our proposed ESD-MBENet network achieves better classification results than previous SOTA networks on RS datasets. In addition, in the field of natural scene image classification, ESD-MBENet also shows strong advantages.
\section*{Acknowledgment}
This work was supported by the National Natural Science Foundation of China [grant number 62072021] and the Fundamental Research Funds for the Central Universities (No. YWF-21-BJ-J-534).

\ifCLASSOPTIONcaptionsoff
  \newpage
\fi


%

\bibliographystyle{IEEEtran}
\bibliography{IEEEfull,mybibfile}




%


\begin{IEEEbiography}[{\includegraphics[width=1in,height=1.25in,clip,keepaspectratio]{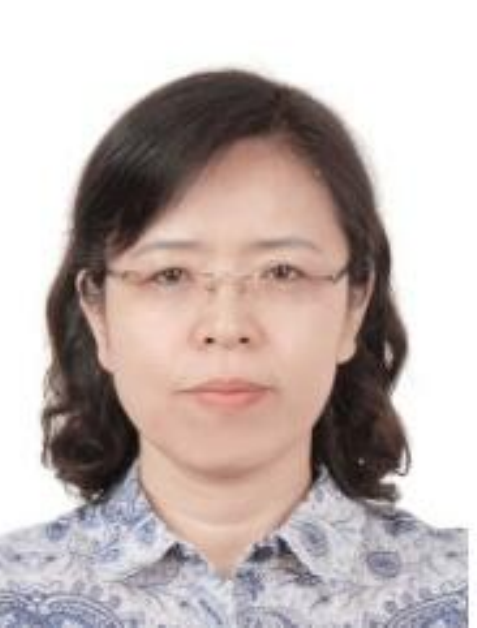}}]{Qi Zhao}
received Ph.D in communication and information system from Beihang University, Beijing, China. She is a professor and works in Beihang University. She was in the Department of Electrical and Computer Engineering at the University of Pittsburgh as a visiting scholar from 2014 to 2015. Since 2016, she has been working on wearable device based first-view image processing and deep learning based image recognition. Her current research interests include one-shot semantic segmentation, communication signal processing and target tracking.
\end{IEEEbiography}
\begin{IEEEbiography}[{\includegraphics[width=1in,height=1.25in,clip,keepaspectratio]{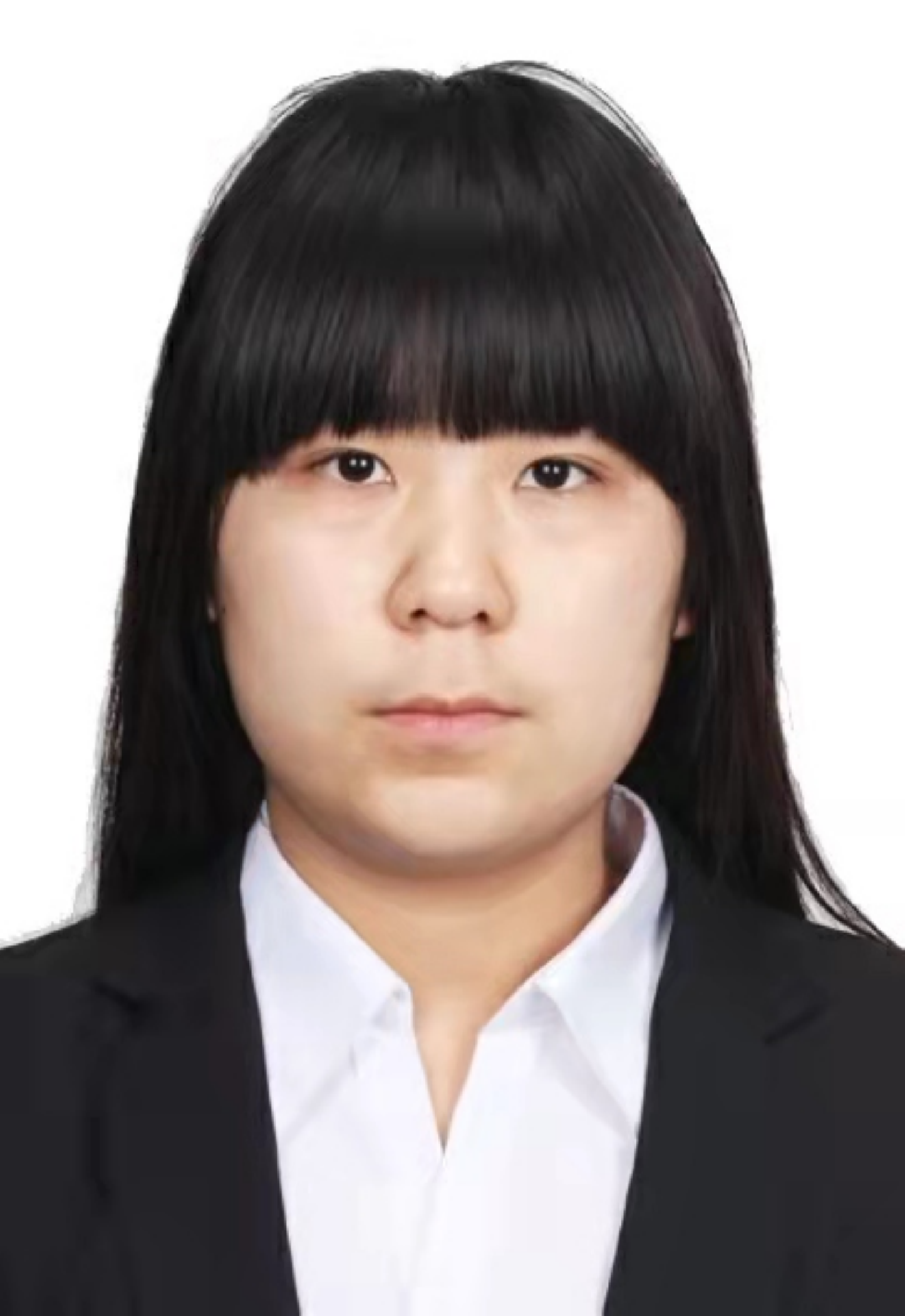}}]{Yujing Ma} received B.S. degree in information engineering from China University of Mining and Technology, Beijing, China, in 2019. She is currently pursuing the M.S. degree with the school of electronics and information engineering, Beihang University, Beijing, China. Her research interests include deep learning,
remote sensing and active learning.
\end{IEEEbiography}

\begin{IEEEbiography}[{\includegraphics[width=1in,height=1.25in,clip,keepaspectratio]{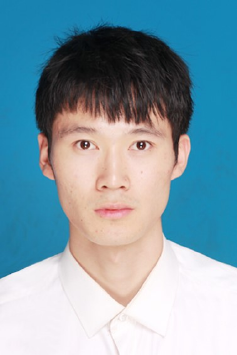}}]{Shuchang Lyu} received the B.S. degree in communication and information from Shanghai University, Shanghai, China, in 2016, and the M.E. degree in communication and information system from the School of Electronic and Information Engineering, Beihang University, Beijing, China, in 2019. He is currently pursuing the Ph.D. degree with the School of Electronic and Information Engineering, Beihang University, Beijing. His research interests include deep learning, image classification, one-shot semantic segmentation and object detection.
\end{IEEEbiography}

\begin{IEEEbiography}[{\includegraphics[width=1in,height=1.25in,clip,keepaspectratio]{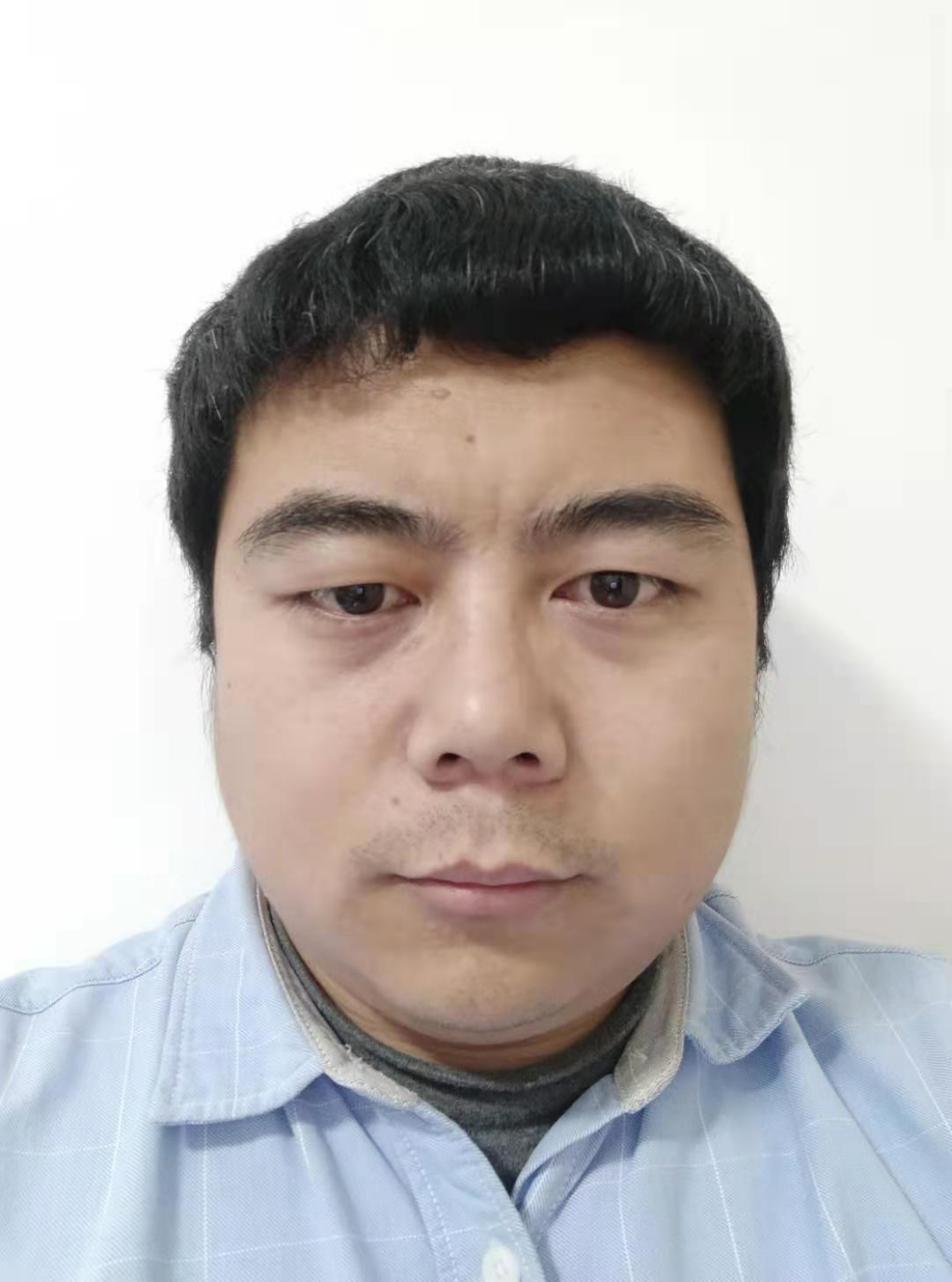}}]{Lijiang Chen}
received B.S. and Ph.D. degrees in the School of Electronic and Information Engineering from Beihang University in 2007 and 2012 respectively. He was a Hong Kong Scholar in the City University of Hong Kong from 2015 to 2017. He is now an assistant professor with the School of Electronic and Information Engineering, Beihang University. His current research interests include pattern recognition, image processing, and human-computer interaction.
\end{IEEEbiography}



\end{document}